\documentclass[conference]{IEEEtran}
\IEEEoverridecommandlockouts
\usepackage{cite}
\usepackage{amsmath,amssymb,amsfonts}

\usepackage{graphicx}
\usepackage{textcomp}
\usepackage{hyperref}
\usepackage{booktabs}
\usepackage{multirow}
\usepackage{array,multirow}
\usepackage[dvipsnames]{xcolor}
\usepackage[T1]{fontenc}\usepackage{tikz}
\usepackage{amsmath, amsfonts}
\usepackage[]{algorithm2e}
\usepackage{algpseudocode}
\def\BibTeX{{\rm B\kern-.05em{\sc i\kern-.025em b}\kern-.08em
    T\kern-.1667em\lower.7ex\hbox{E}\kern-.125emX}}

\newcolumntype{R}[1]{>{\raggedleft\arraybackslash }b{#1}}
\newcolumntype{L}[1]{>{\raggedright\arraybackslash }b{#1}}
\newcolumntype{C}[1]{>{\centering\arraybackslash }b{#1}}
\usetikzlibrary{arrows.meta} 

\def\nbTasks{\ 3 } 
\def\nbDataSets{\ 6 } 
\begin{document}

\setlength{\arrayrulewidth}{0.4mm}
\renewcommand{\arraystretch}{2}
 \makeatletter
    \newcommand{\linebreakand}{%
      \end{@IEEEauthorhalign}
      \hfill\mbox{}\par
      \mbox{}\hfill\begin{@IEEEauthorhalign}
    }
    \makeatother

\title{
Modality Influence in Multimodal Machine Learning
\thanks{This research is performed under the MESRS Project  PRFU: C00L07N030120220002 }
}

\author{\IEEEauthorblockN{Abdelhamid Haouhat}
\IEEEauthorblockA{\textit{Lab. d'Informatique et de Mathématiques} \\
\textit{Université Amar Telidji}\\
Laghouat, Alg\'erie  \\
a.haouhat@lagh-univ.dz}
\and
\IEEEauthorblockN{Slimane Bellaouar}
\IEEEauthorblockA{\textit{Lab. d'Informatique et de Mathématiques} \\
\textit{Université de Ghardaïa}\\
Ghardaïa, Alg\'erie  \\
bellaouar.slimane@univ-ghardaia.dz}
\linebreakand
\IEEEauthorblockN{Attia Nehar}
\IEEEauthorblockA{\textit{Lab. d'Informatique et de Mathématiques} \\
\textit{Université Ziane Achour}\\
Djelfa, Alg\'erie  \\
a.nehar@lagh-univ.dz}
\and
\IEEEauthorblockN{Hadda Cherroun}
\IEEEauthorblockA{\textit{Lab. d'Informatique et de Mathématiques} \\
\textit{Université Amar Telidji}\\
Laghouat, Alg\'erie \\
hadda\_cherroun@lagh-univ.dz}

}

\maketitle

\begin{abstract}
Multi-modal Machine Learning has emerged as a prominent research direction across various applications such as Sentiment Analysis, Emotion Recognition, Machine Translation, Hate Speech Recognition, and Movie Genre Classification. This approach has shown promising results by utilizing modern deep learning architectures. Despite the achievements made, challenges remain in data representation, alignment techniques, reasoning, generation, and quantification within multi-modal learning. Additionally, assumptions about the dominant role of textual modality in decision-making have been made. However, limited investigations have been conducted on the influence of different modalities in Multi-modal Machine Learning systems.
This paper aims to address this gap by studying the impact of each modality on multi-modal learning tasks. The research focuses on verifying presumptions and gaining insights into the usage of different modalities. The main contribution of this work is the proposal of a methodology to determine the effect of each modality on several Multi-modal Machine Learning models and datasets from various tasks. Specifically, the study examines Multi-modal Sentiment Analysis, Multi-modal Emotion Recognition, Multi-modal Hate Speech Recognition, and Multi-modal Disease Detection.
The study's objectives include training state-of-the-art Multi-Modal Machine Learning models with masked modalities to evaluate their impact on performance. Furthermore, the research aims to identify the most influential modality or set of modalities for each task, and draw conclusions for diverse multi-modal classification tasks. By undertaking these investigations, this research contributes to a better understanding of the role of individual modalities in multi-modal learning and provides valuable insights for future advancements in this field.
\end{abstract}


\section{Introduction}

Multi-modal Machine Learning has recently become a centric research direction in many  applications. One can mention  Sentiment Analysis,   Emotion  Recognition~\cite{jiang2021multitask,fortin2019multimodal,du2022gated,xu2022multimodal,arano2021multimodal}, Machine Translation\cite{elliott2017imagination,gronroos2018memad, yao2020multimodal}, Hate Speech Recognition~\cite{kiela2020hateful,Gomez_2020_WACV,yang-etal-2019-exploring-deep}, Movie Genre classification, nd achieve great results.

In addition, multi-modal learning  has involved  important research interest and overcome while  using different modern deep learning architecture~\cite{vaswani2017attention,lecun1995convolutional}.

Despite these outstanding achievements with various strategies and models, multi-modality' superiority to uni-modality performance has been  demonstrated both empirically and theoretically\cite{huang2021makes,tsai2020multimodal,poria-etal-2015,xu2022multimodal}.
Nonetheless, there are several difficulties with multi-modal learning due to the  data representation (e.g. representation of modalities and their  fusion) and alignment techniques (e.g. Correspondence of modal  elements), reasoning, generation, and quantification.  Morency et al.~\cite{morency2022tutorial} brought more insights on those challenges.


 Some works show that   multi-modality is not mostly required to achieve better results in all data by dealing in the same ways, especially with easy and simple samples\cite{xue2022dynamic}.

Furthermore, many assumptions are being made about "textual modality" that it  is the dominant modality for participating and making the true decision when combined with other modalities~\cite{kiela2019supervised, kiela2019supervised}.

However,  very   few investigations have been drawn on  modality influence in  Multi-modal Machine Learning based systems. In this paper, we will emphasize the influence of each modality on multi-modal learning tasks in order to study and verify some of the  aforementioned presumptions and give more insightS on different modalities usage.


The main contribution of this work is to propose a  methodology to determine the effect of each modality within several Multimodal Machine learning  and on different  datasets from various tasks: Multi-modal Sentiment Analysis MSA, 
, Multi-modal Emotion Recognition, 
Multi-modal Hate Speech recognition,
Multi-modal Disease Detection and 

The study focuses: 
\begin{enumerate}
    \item On training the state-of-the-art models of Multi-Modal Machine Learning  with masking some modalities and evaluating their  impact on performances.
   \item  For each task, we intend to squeeze out which modality (or set of modalities)is more impactful on the learner's performances 
   \item On drawing conclusions for  diverse set of multi-modal classification tasks.
\end{enumerate}
  
The remainder of the paper is organized as follows. Section~\ref{sec:BG} gives a glance at the machine learning methods deployed in Multi-modality in addition to fusion techniques. Section~\ref{sec:RW} provides an overview of previous research on multi-modal tasks  and  the impacts of modalities.   In Section~\ref{sec:Methodology}, we describe the followed methodology while presenting the built ML models. Section~\ref{sec:Experiments} is dedicated to describe the batch of made  experiments,  the interpretation of the  results and the empirical findings. 
Finally, Section~\ref{sec:conclusions} outlines the conclusions and future works.

\section{Background}
\label{sec:BG}

In this paper, we focus on state-of-the-art MLL  models based on deep learning, we inspired from \cite{ma2022multimodal} and many more work  to finetune  a pretrained~\cite{pmlr-v139-radford21a} CLIP,\cite{devlin2018bert}  BERT,~\cite{ pennington2014glove} GloVe, multimodal bi-transformer model (MMBT) models,

\subsection{Pre-trained models} 
 \subsubsection*{GloVe } One of the famous word embedding models due to  Pennington et al.~\cite{pennington2014glove}. It  is based on an unsupervised learning algorithm for obtaining vector representations for words. Training is performed on aggregated global word-word co-occurrence statistics from  corpora.  
  \subsubsection*{BERT} Bidirectional Encoder Representations from Transformers  developed by Google~\cite{devlin2018bert}. BERT was created and published in $2018$ by Jacob Devlin and his colleagues from Google.  BERT large encompasses  $24$ encoders with $16$ bidirectional self-attention heads trained from unlabeled data extracted from the BooksCorpus with $800$M words and English Wikipedia with $2,500$M words.
\subsubsection*{CLIP}  Contrastive Language-Image Pre-training developed by OpenAi in $2021$, CLIP has separated transformer encoders for text and image modalities to image-text, zero-shot classifications~\cite{pmlr-v139-radford21a}.
  
\subsubsection*{MMBT}  
Facebook introduces a supervised multimodal bitransformer model that jointly finetunes unimodally pretrained text and image encoders
by projecting image embeddings to text token space  and match state-of-the-art accuracy on several 
classification tasks 
for multimodal BERT-like architectures\cite{kiela2019supervised}.

\subsection{Fusion Methods}
 Basically, We have two main methods to make a fusion of modalities the first is Early fusion, which happens when we mix the modalities before making decisions with concatenation, summation, or cross-attention mechanism, then we have a late fusion method by doing a prediction based on each modality  alone then combine the decision to get a final prediction\cite{xu2022multimodal}.

Assume we have $m_{1}, m_{2}, ..., m_{n} \in \mathbb{Z} $ , where $n $ is the number of modalities and $ \mathbb{Z} $ is the modalities Feature space.
 suppose   $  \phi (.)  $  is the fusion function and takes as input the modalities $M $ , we can define it  with these two methods in \
\subsubsection*{Concatenation or summation }
 
\begin{equation}
 \phi (M) = (concat(m_{1}, · · · , m_{n})  
  \parallel sum(m_{1}, · · · , m_{n})) \odot  W 
\end{equation}

\begin{figure}
    \centering
    \includegraphics[scale=0.5]{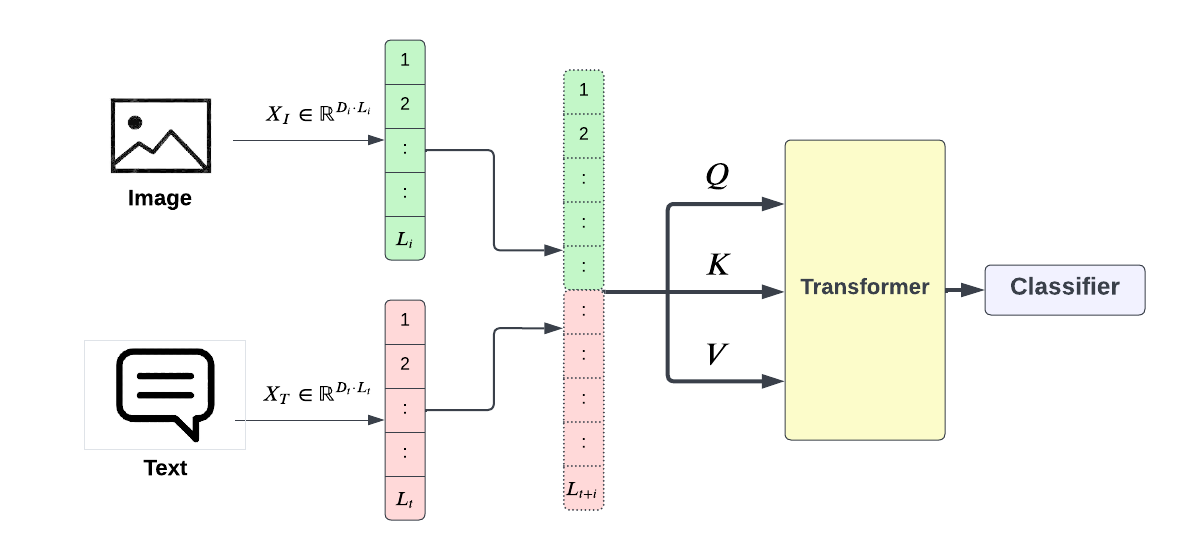}
    \caption{Fusion modalities by Concatenation}
    \label{fig:concat}
\end{figure}
\subsubsection*{Cross Attention }

\begin{equation}
 \phi (m_{x},m_{y}) =    Softmax(  \frac{m_{1}W_{1}^{q} \odot  m_{n}(W_{n}^{k})^{T}}{\sqrt{d_{q}}})  m_{n}W_{n}^{v}     
\end{equation}
\begin{figure}
    \centering
    \includegraphics[scale=0.5]{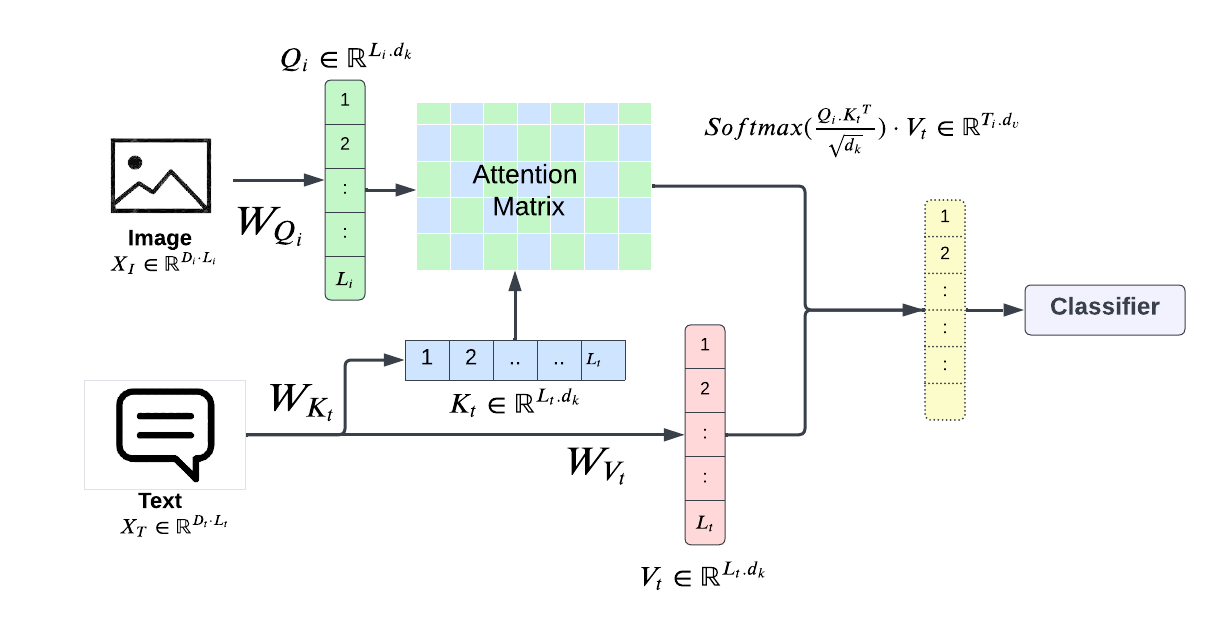}
    \caption{Fusion modalities with CrossAttention}
    \label{fig:cross}
\end{figure}
Figures \ref{fig:concat},\ref{fig:cross} illustrates the two methods,
in this paper, we use early fuse approaches 

\section{Related Work}
\label{sec:RW}


 In our study, we review the related work from broad to narrow. In fact, we start by presenting the multi-modal learning works in general, then we deal with specific works that focus on the influence of multi-modality on machine learning. 
 
Multi-modal modeling implies fixing  mainly three  aspects: the deployed deep Learning method for each category of modality,  the technique of   integration of these modalities' data,  and the techniques  of the fusion.

\subsection{Multi-modal learning}
Multi-modality has become prevalent in various Machine Learning tasks such as sentiment analysis, emotion recognition, translation, medical classification.  Multi-modality has also found applications in various other areas such as image classification, speech recognition, natural language processing, and recommendation systems, among others.

One of the earlier works in multimodal deep learning is the work of Ngiam et al. \cite{DBLP:conf/icml/NgiamKKNLN11}. The authors proposed  deep architectures to learn features over multiple modalities. In fact, they are interested in discovering the correlations between audio and visual data for speech recognition ("mid-level" relationships). They trained and tested their proposed deep autoencoders on audio-visual classification of isolated letters and digits using the available datasets (CUAVE, AVLetters, AVLetters2, Stanford Dataset, TIMIT). The obtained experimental results reveal that the correlation between multiple modalities is still challenging and they show how deep learning can be applied for discovering multimodal features. 




For  the sentiment analysis task,  Poria et al.~\cite{poria-etal-2015} use three modalities. The image features are extracted with a CNN model concatenated with textual Word2Vec embeddings and audio features. They  investigate a  variety of  fusion techniques namely  early feature  fusion techniques and late prediction    fusion ones. They demonstrate that fusion modalities methods are crucial in MML. 


Another intriguing attempt at RGB-D object recognition, Where Eitel et al \cite{eitel2015multimodal} aims to combine  SOTA deep learning models CNNs and MLP in this task, Actually, the proposed architecture concatenates the two learned features from two separate pre-trained CNNs one for color RGB image and the second for depth image modality, they integrate these two with a simple multi Layer perceptron in a late fusion approach, while the  training of these networks follows a stage-wise approach for  more accurate learning, Authors recommend a data augmentation scheme in case of data perturbation  (e.g. a noisy data acquired by an imprecise sensor). However, They finetune RestNet Features for depth image in this approach and achieve better results but they have to do extra work in order to tackle the lack of labeled depth data datasets, 
According to the emergence of the attention mechanism\cite{bahdanau2014neural},  Huang et al\cite{huang2016attention} proposes to parallel long short-term memory (LSTM) threads and generate representation features by concatenating dispersed global and regional visual data with text features and then make it feasible to attend while the decoding and  generate expected  target sequences, authors used the attention based decoder to defeat the missing semantic context and That can assist the decoder to guess the following word by forming the context. 
Due to  the  Stunning success of the attention mechanism and its use in Transformers-based architecture  in Natural Language Processing\cite{vaswani2017attention} and Vision Computers\cite{dosovitskiy2020image},  Researchers interested in a  promising  research direction, which has greatly promoted the development of  various MML tasks and overcoming its challenges (e.g., Fusion, Representation, Alignment, Robustness, Efficiency .. etc)\cite{xu2022multimodal} using these stacks.
Keith et al, \cite{arano2021multimodal} propose a new hyperbolic model that overcoming one of the main issues (i.e. the input representation space) in MML and get a rich data representation through mapping the input features from the original Euclidean input space to a hyperbolic space in order to capture the hierarchical structure of the relations among the input elements and increases prediction performances. They also investigate how small and large networks impact the results, their experimentation also showed superior performance if combine hyperbolic and Euclidean layers in the model. 
Gabeur et al, \cite{gabeur2020multi} aims to tackle the task of caption-to-video by a cross-modal framework of two deep learning encoders BERT~\cite{devlin2018bert} for the caption and MMT for video then estimate the similarity between a source and target candidates(i.e. source and target may be a caption or video), They thus exploit the cross-modal concept to let the  video constituent modalities interact with each other to get a higher-level semantics video representation  from the lens  of multi-modality and the temporal information  in videos, This last showed an important factor for outperform prior SOTA in three  benchmark datasets.
Text-Image are very widespread, particularly in product reviews, Yongping et al \cite{du2022gated} believe that the fine-grained features and the attributes extracted from an image have affected the performances regarding its use and its quality, Therefore leading authors to  adopt a gated attention mechanism to fuse the textual features with image features extracted by pre-trained convectional neural network models, Also by introduce this gated attention mechanism it helps to reduces the image noise affect to decide where the model ignore the unsolicited image features. Since researchers leverage Transformer-based  architectures to prevent the alignment problem of multimodal data For instance,  They address the synchronization of video-audio sequences across time and the proper audio correspondence to video, Consequently,  Morgado et Vasconcelos \cite{morgado2020learning}  treat the alignment problem through the correlation spatial information in audio-visual content and propose contrastive networks that perform audio-visual spatial alignment of 360$^{\circ}$video using a transformer architecture to combine representations from multiple viewing angles. However, 360$^{\circ}$video is still less prevalent than regular video affecting the quality of representation learned  but this specific data creation is likely to grow substantially in the future, 
Nowadays, multimodal machine learning has attracted more and more attention \cite{paul2022vision,ma2022multimodal,hazarika2022analyzing}  to evaluate how the Transformer components contribute  to the robustness and the performance in many downstream tasks.

\subsection{ Modality influence }
There are several attempts to handle the missing modality and its influence on multimodal learning. Ma et al \cite{ma2022multimodal} investigate the influence of missing modality in  transformers-based architecture and find out that  Transformer models are sensitive to missing modalities while different modal fusion strategies will significantly affect the robustness against this missing, they propose a  method  by automatically searching for an optimal fusion strategy regarding input data and show superior performance when dealing with incomplete data. 
The encoder-based approaches are  one of the proposed solutions to almost missing and incomplete data problems in machine learning. Therefore, Ma et al\cite{ma2021smil}  address the problem of missing modality by proposing a new model, severely missing modality (SMIL) treats missing modalities in terms of flexibility (i.e. missing  in training, testing, or both), for that SMIL reconstructs the missing Modality by a trained network on a modality-complete dataset, and approximates weights of the priors learned instead of directly generating the missing modality. They demonstrate consistent efficiency  across different benchmarks, MM-IMDb, CMU-MOSI, and avMNIST.
Besides, Empirical Research  treatments in this new influential work direction, Huang et al\cite{huang2021makes} prove theoretically that using all available modalities is  better performance in terms of population risk than using only a subset of data through most  multi-modal fusion methods and highlight the conditions to get better latent space. 
Valentin et al ~\cite{gabeur2022masking} address the unobserved modality drawback when using some pre-training methods via  training a video encoder, The encoder follows an alternating
modality masking strategy, They alternate video modalities(i.e. RGB, audio, and ASR transcripts) to be masked and predicted it with a supervising of the other available modalities, the model implemented using the MultiModal Transformer(MMT) with shared parameters across all layers this makes the model to optimize the parameters even when one of them are masked, Finally, They achieve improved performance  for video retrieval task on the How2R, YouCook2, and Condensed Movies datasets.
These previous works focus on how MML deals with noisy or missing modalities, they  proposed solutions to enhance  models and designed architectures in ways to deal with imperfect data. In contrast, our  proposed methodology is to measure the real impact of each modality's  participation  in the downstream task.

\section{Methodology}
\label{sec:Methodology}

In order to gain a deeper understanding of how modalities affect the performance of multimodal Learning models, we conducted empirical investigations as part of our assessment. based on three main factors: Variety of targeted tasks, variety in datasets, and the chosen Deep-learning models. 
For the first factor,  we have  targeted a set $T$ of  \nbTasks different tasks. These tasks include Multimodal Sentiment Analysis (M-SA), Multimodal Emotion Recognition (M-ER), Multimodal Hate Speech Recognition (M-HSR), and Visual Question Answering (QA). 
In fact,  to get more significant results, we have varied the scope of tasks among those most used in real-life applications.

Concerning the testbed, and for the sake of obtaining results that are more independent of data,   we have chosen  more than two datasets, when possible,  for each targeted task.  This is important for the study to achieve trustworthy results.  

Regarding  the third factor, we have tried to choose and build multimodal models that are among the state-of-the-art and based on mainly Deep-Learning.

\begin{figure}
    \centering
    \includegraphics[scale=0.2]{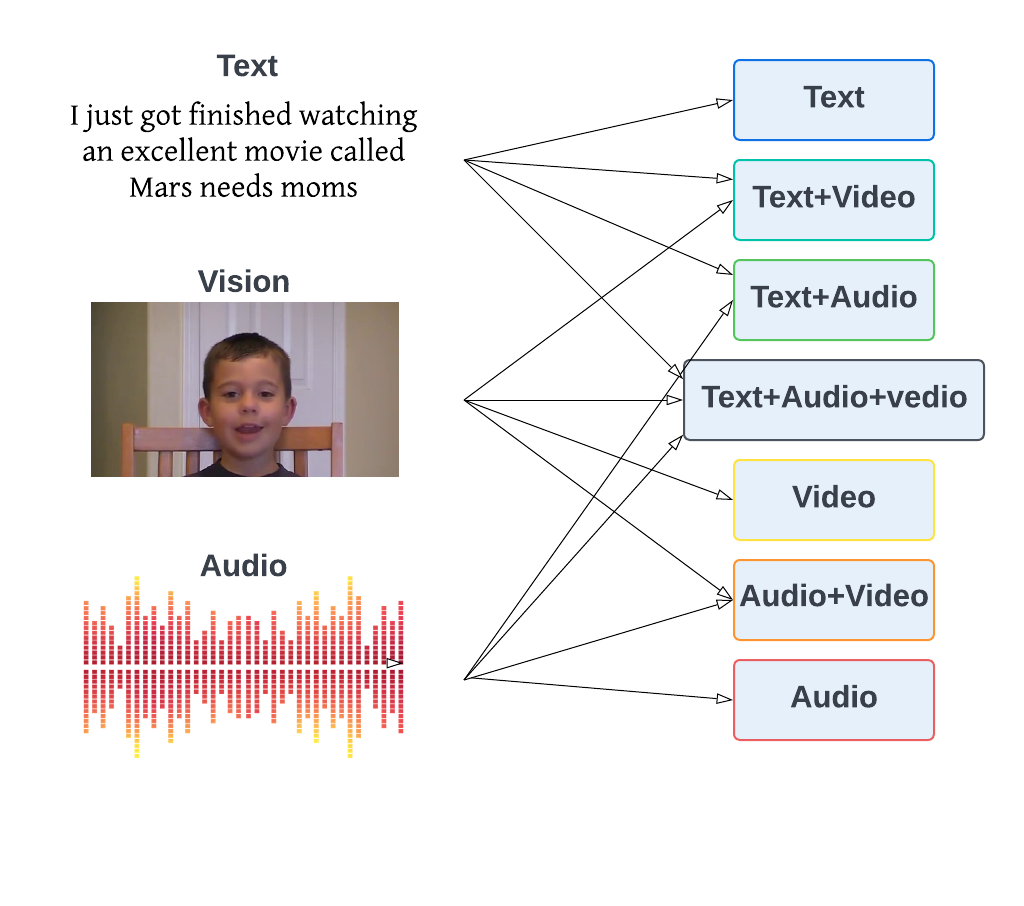}
    \caption{Different Built Models for Task$_i$}
    \label{fig:built-model}
\end{figure}

For each task$_i \in T$,  we generate  eight models according to the available  modalities in the datasets. These models are unimodal ones where only one modality is considered. The  bi-models are those which dealt two modalities.  and the three-Modals.   Figure~\ref{fig:built-models} illustrate the diff built models needed for the evaluation.

\RestyleAlgo{ruled}
 \begin{algorithm}
\caption{Our methodology algorithm}

\SetKwInput{KwData}{Tasks}
 \KwData{ $ task_1$;  $ task_2$  ;  $ task_3$  ;  $ task_4$ ; ... ; $ task_i$. }
 \SetKwInput{KwData}{datasets}
 \KwData{$ dataset_1$,  $ dataset_2$  ,  $ dataset_3$  , ... , $ dataset_j$ .}
 
 \SetKwInput{KwData}{Tasks}
  \For{ \textbf{each}    $ \top  $ from \KwData }{
  \SetKwInput{KwData}{datasets}
  define  $ \partial_j$ as a database variable;
  
 \For{ \textbf{each}    $ \partial_j$ belong to  $\top  $ }{
 define  $ \kappa  $ as the number of all modalities in  $ \partial_j$;

  \For{ sub\_set S\{m\} \textbf{from} $ m=1    \mapsto  2^{\kappa} -1 $ }{

  choose $ \theta_{m} $ an appropriate \textbf{model} \;
  where $\{ \theta_{\kappa+1}  ..  \theta_{2^{\kappa} -1} \}$  multimodal models;
  and   $ \{\theta_{1} , ...,  \theta_{\kappa}\} $ are unimodal models \;

 1)\
  Using S\{m\}  train $ \theta_m $ get  experiments  \;
    2)
      save the results with the same metric performance

  }
  Evaluate obtained results  and make the final observation 
  
  }
  
 }

\end{algorithm}
In what follow we will describe the deployed model per task.
\subsection{Multi-model Sentiment Analysis task}
The study of sentiment analysis is becoming more and more significant in both academia and industry. So, the Multimodal sentiment analysis's goal is to determine a person's attitude's polarity  from all available modalities (text, audio, video, etc.). The multimodal technique improves sentiment analysis performance, as shown by academic studies and empirical research \cite{soleymani2017survey}.
  
For the generation of the multimodal Sentiment Analysis task, we have used a model inspired by the work of~\cite{tsai2019multimodal}, we use the model illustrated in Figure \ref{fig:multi}
by feeding the embeddings of all modalities to six crossmodal modules to fuse features of all pairs of modalities.  
\begin{figure}
    \centering
    \includegraphics[scale=0.4]{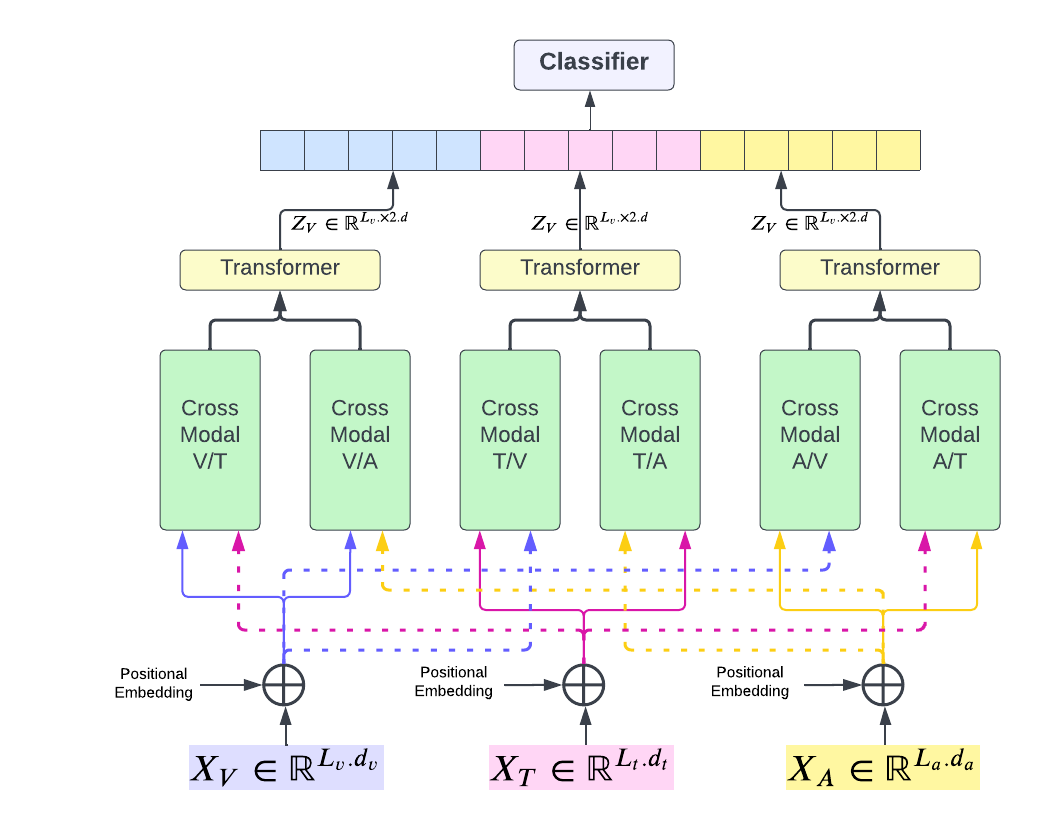}
    \caption{Sentiment analysis and emotion recognition model}
    \label{fig:multi}
\end{figure}

In 
For unimodal we eliminate the two other modalities and feed the features of specific modality with self-attention.
\subsection{Multimodal Emotions Recognition  Model}
Human emotions are expressed in a variety of ways, including facial expressions and audible sounds, and also humans perceive the world in a multimodal way. 
Multimodal Emotions identify human emotion by taking not only text or video but all features from facial expressions, spoken expressions, written expressions. 

For the generation of the multimodal  Emotion recognition  task, we have used the same model in the sentiment analysis task referred to above Figure\ref{fig:multi}.

\subsection{Multimodal Hate Speech Model}
Hate speech is defined as \href{ https://www.facebook.com/notes/344897590097197/}{(Facebook, 2016},\href{ https://help.twitter.com/en/rules-and-policies/hateful-conduct-policy}{ Twitter, 2016)}:

“Direct and serious attacks on any protected category of people based on their race, ethnicity, national origin, religion, sex, gender, disability or disease.“

The multimodal Hate Speech (MHS) model takes as input all modalities such as Image Text or Audio to a designed architecture or pre-trained models to detect if it has a hateful speech or not \cite{gomez2020exploring}. 

In this task, we have used a model inspired by the work Rostyslav Neskorozhenyi. Figure\ref{ref:hatespeech} illustrates the deployed model.
\begin{figure}
    \centering
    \includegraphics[scale=0.21]{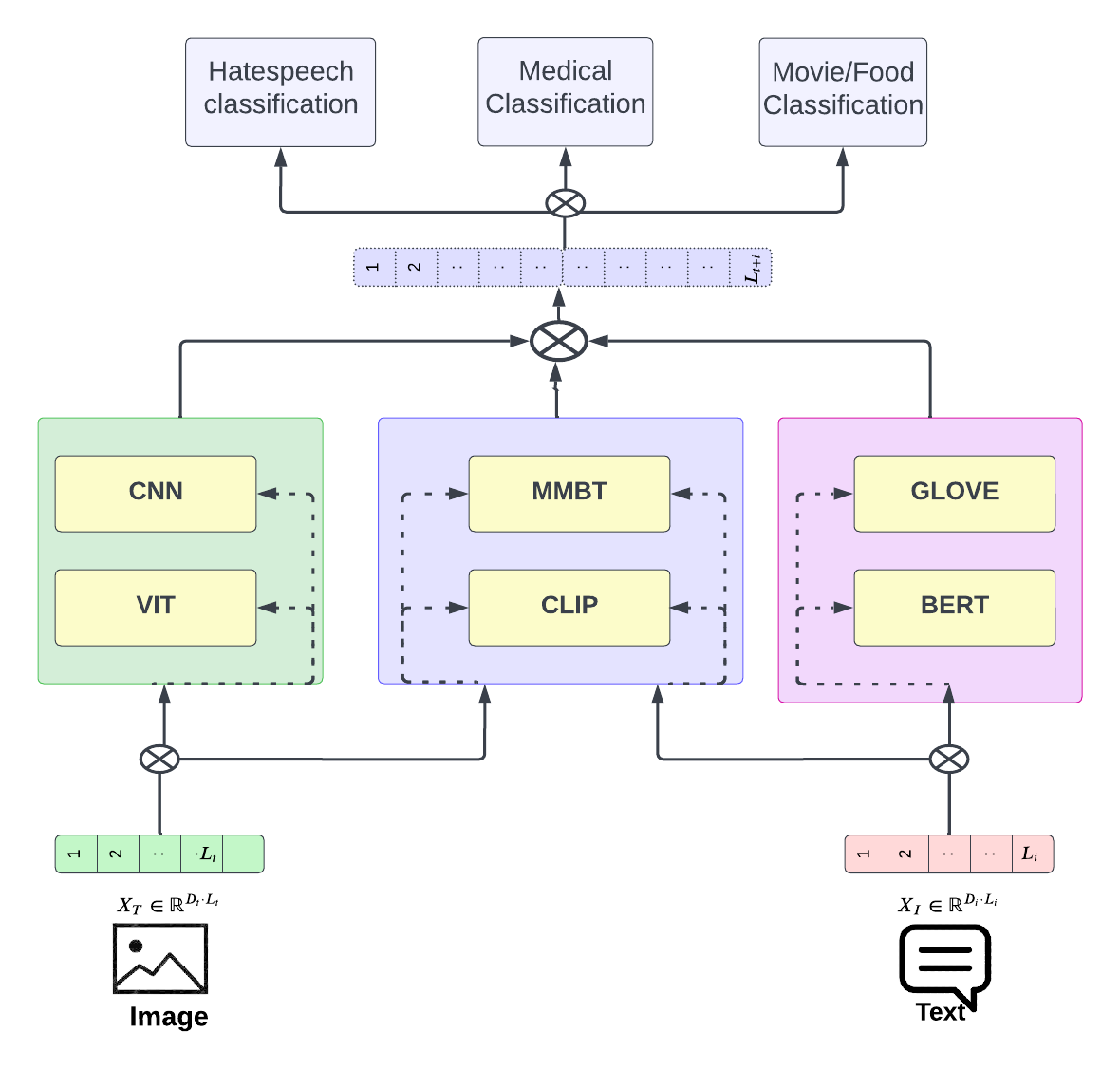}
    \caption{Sentiment analysis and emotion recognition model}
    \label{fig:hatespeech}
\end{figure}

\subsection{Multimodal Machine Translation}
Multimodal machine translation (MMT) aims to create better translation systems by integrating the textual context with a second modality, typically pictures.

In this task, we will use the same architecture of two modalities when we extract textual features from GolVe model and on another hand the image features extracted from the last convolutional layer of ResNet-50.    
In the multimodal case, we feed these two feature vectors to a standard transformer to get results and generate translation.
In unimodal cases, We use a transformer for both textual and visual features even in the case of the image only we incorporate
features  as words in the source sentence, this technique is used in \cite{calixto2017incorporating}.
\subsection{Medical  }

\section{Experiments}
\label{sec:Experiments}
Our codes and details on the used
datasets are available{\url{http://github.com/belgats/Modalitites-Impact-In-ML}}

\subsection{Experiment Setup and Used Metrics }
\label{ssec:Experiment-Setup}


\subsubsection*{Datasets}
 We have deployed \nbDataSets 
 
 In our experimental settings we analysed the CMU-MOSEI~\cite{zadeh2018multimodal}, CMU-MOSI~\cite{zadeh2016mosi}, UR-FUNNY\cite{hasan2019ur } , Hate speech memes~\cite{kiela2020hateful } datasets. 
 
 \textbf{CMU-MOSEI} is a collection of in-the-wild videos annotated with both sentiments and emotions. It contains more than 65 h and 23,500 utterances of 3228 videos from more than 1000 online YouTube speakers and 250 topics. Sentiments are labeled with values in the range, which are binned into  \textbf{two} (negative, positive),  \textbf{five} (negative, weakly negative, neutral, weakly positive, positive), and  \textbf{seven} (strongly negative, negative, weakly negative, neutral, weakly positive, strongly positive) sentiment classes. 
 
 \textbf{CMU-MOSI}  It contains 2 h and 2199 sentiment segments of 93 videos from 98 distinct speakers. This dataset has the same sentiment annotations as in CMU-MOSEI.
  On the other hand, emotions are annotated with six labels that correspond to the basic emotional states proposed in (Ekman, 1999): happy, anger, sad, disgust, fear, and surprise. Specifically, the presence and intensity of a given emotion $e$ is rated on a Likert scale of 1–3 as follows: 0 for no evidence of $e$, 1 for weak $e$, 2 for $e$ and 3 for high $e$. 
  
 \textbf{IEMOCAP}~\cite{busso2008iemocap }
  the Interactive Emotional Dyadic Motion Capture (IEMOCAP) database, which contains three modalities, Text, Video, and Audio. We follow the data preprocessing method of [33] and obtain 100 dimensions of data for audio, 100 dimensions for text, and 500 dimensions for video. There are six labels here, namely, happy, sad, neutral, angry, excited, and frustrated.
  
 \textbf{UR-FUNNY}~\cite{hasan2019ur } collection of 1866 videos with transcription from TED Portal  for understanding humor, which has three modalities of features extracted from Text, Visual, and Audio as follows: a pre-trained Glove word embedding for the textual modality, COVAREP software is used to extract the Acoustic features at a rate of 30frames/sec, and OpenFace facial behavior analysis tool is used to extract facial expression features.

  \textbf{Multi30k}   
~\cite{elliott2016multi30k} consists of two multilingual expansions of the original Flickr30k dataset referred to as M30kT and M30KC, respectively. Multi30k contains 30k images, and for each of the images, M30kT has one of its English descriptions manually translated into German by a professional translator. M30KC has five English descriptions and five German descriptions, but the German descriptions were crowd-sourced independently from their English versions. The training, validation, test sets of Multi30k contain 29k, 1014, and 1k instances respectively.

 \textbf{MM-IMDB}~\cite{hasan2019ur } 
  is the largest multimodal dataset for genre  classification movies, It comprises 25,959 movie titles, metadata, and movie posters. The task is to perform multi-label classification of 27 movie genres from posters (image modality) and text descriptions (text modality). We follow the original data split in ~\cite{arevalo2017gated } and use 15,552 data for training, $2,608$ for validation, and $7,799$ for testing
 
\textbf{MMHS150K}~\cite{Gomez_2020_WACV }
   formed by 150,000 tweets, each one of them containing text and an image, classed into  6 categories: No attacks to any community, racist, sexist, homophobic, religion-based attacks, or attacks to other communities.

\textbf{Food-101}~\cite{Gomez_2020_WACV }
 consists of 101 food categories with 750 training and 250 test images per category, making a total of 101k images. The labels for the test images have been manually cleaned, while the training set contains some noise.
\subsubsection*{Used Metrics}
Depending on the used tasks, we evaluate the performance of our Models  on the following standard metrics: F1 score, Accuracy, Mean error, A ...
 


\subsection{Multimodal Sentiment Analysis }
\label{ssec:Experiment-SA}
We present our experimental results on MultiModal Emotion recognition  using the Multimodal transformers inspired by\cite{ tsai2019multimodal}  results in the Tables\ref{tab:M-ER-Performances},  according to Accuracy and F1 score.

 \begin{table*}[t]
		\caption{Performances of SA Models.}
				\begin{center}
			\begin{tabular}{|c|c|C{0.6cm}|C{0.6cm}|C{0.6cm}|C{0.6cm}|C{0.6cm}|C{0.6cm}|C{0.6cm}|C{0.6cm}|}
				\cline{3-9}
                \multicolumn{2}{c|}{}&\multicolumn{7}{c|}{Modalities}\\\cline{3-9}
				\multicolumn{2}{c|}{}		 &T	   &  A   &  V      & TA    & TV & AV &  TVA  \\\hline 
				 \multirow{4}{*}{CMU-MOSEI}&  Accu & 80.43    & 68.70   &  66.80  & 80.90    &    80.71  &  0.6623  &80.71\\ 
				  &  F1 &   79.47   &65.82     & 65.79   & 80.73    &  80.54    &    0.7079 &80.15\\
      & Accu.  &   0.8102 &  0.6555   &  0.6497  &   0.8011  &0.8019    &0.6726  &0.813 \\ 
				  & F1  & 0.8133 & 0.7096 &  0.7150   & 0.8030    &  0.8042     & 0.6987  &0.815 \\\hline
				 \multirow{4}{*}{CMU-MOSI}& Accu  &  0.791    &  0.55   &  0.5823 & 0.80    &      0.759 & 0.5625 & 0.791 \\ 
				  & F1  & 0.79     &   0.561  &  0.579  &  0.80   &   0.757   & 0.573  &0.792 \\
                  & Accu  & 0.7759     &  0.5625   & 0.4786   &   0.8018  & 0.7728   &  0.6021& 0.798 \\ 
				  & F1  & 0.7749      &   0.5746  & 0.5506    &   0.8024  &    0.7714   &  0.6006 &0.801 \\\hline 
      	 
			\end{tabular}
		\end{center}

		\label{tab:M-SA-Performances}
\end{table*}
 
\subsection{Multimodal Emotion recognition   }
\label{ssec:Experiment-ER}
We present our experimental results on MultiModal Emotion recognition  using the Multimodal transformers inspired by\cite{ tsai2019multimodal}  results in the Tables\ref{tab:M-ER-Performances}, according to Accuracy and F1 score.
\begin{table*}[!t]
    \centering
    \begin{center}
			\begin{tabular}{|c|c|C{0.6cm}|C{0.6cm}|C{0.6cm}|C{0.6cm}|C{0.6cm}|C{0.6cm}|C{0.6cm}|C{0.6cm}|}
				\cline{3-9}
                \multicolumn{2}{c|}{}&\multicolumn{7}{c|}{Modalities}\\\cline{3-9}
				\multicolumn{2}{c|}{}		 &T	   &  A   &  V      & TA    & TV & AV &  TVA  \\\hline 
				 \multirow{4}{*}{IEMOCAP}  & F1  &  0.646&  0.659 & 0.646  & 0.64  & 0.65  & 0.649  &0.66 \\ 
			    	  &    Accu  &     0.749 &   0.752   &   0.749  & 0.749    &   0.74    &0.74   & 0.76 \\
          & F1  &  0.7654 &  0.7903  &  0.6776 & 0.7871  &  0.7655 &  0.7778  & 0.8087\\ 
			    	  &    Accu  &  0.7750     & 0.7985     &  0.7515    &      0.7960 &   0.7769     &  0.7851 & 0.8163  \\
				  \hline
				   \multirow{2}{*}{UR-FUNNY}  & F1  & 0.671 &  0.569  &  0.563  & 0.681  &   0.666&  0.566  &0.679 \\ 
			    	  &    Accu  & 0.668  &  0.568 &   0.553 &   0.679   &  0.658    &  0.56 & 0.678 \\ 
          
				  \hline
			\end{tabular}
\end{center}

    \caption{Details on the   IEMOCAP Dataset for  Emotion Recognition}
    \label{tab:M-ER-Performances}
\end{table*}

 \subsection{MultiModal Hate Speech}

We present our experimental results on MultiModal Hate Speech detections using the TextBERT model for treating the textual modality,   and for the Image modality we  use the Vision Transformer (Vit) model to extract features and make predictions, finally, we use VisualBERT model for both modalities, results in the Table\ref{tab:M-HS-Performances}  according to BLEU and METEOR which used in prior works.
 \begin{table}[ht]
		\caption{Performances of Hate Speech task on Hate Memes and MMHS150K datasets.}
				\begin{center}
			\begin{tabular}{|c|c|C{0.6cm}|C{0.6cm}|C{0.6cm}|}
				\cline{3-5}
                \multicolumn{2}{c|}{}&\multicolumn{3}{c|}{Modalities}\\\cline{3-5}
				\multicolumn{2}{c|}{} &  T  & I      & TI       \\\hline 
               \multirow{2}{*}{ Hate Memes}  &  Accu     &    59.2  &  55.92  &  68.20  \\  
			  & AUROC    &   65.08   &  52.13   & 74.75    \\\hline		
               \multirow{2}{*}{ MMHS150K    }     &  Accu     &   67.4  & 56.8  &  68.4 \\ 
			  & AUROC    &  72.3  &    58.6    & 73.4   \\\hline	
     
			\end{tabular}
		\end{center}

  
  
		\label{tab:M-HS-Performances}
\end{table}

 \subsection{MultiModal Translation  }
 \label{ssec:MultiModal-translation }
We present our experimental results on the German-English dataset using the Word2Vec  model pretrained on the textual modal only  and using only  Image modality only with the VGG Network model to extract features and make predictions, finally, we use  MMT model~\cite{yao2020multimodal} for both modalities, results in the Table~ according to BLEU and METEOR which used in prior works.

\begin{table}[]
    \centering
    \begin{tabular}{L{1cm}L{2cm}L{1.6cm}L{2.4cm}}
\toprule

 Mertic &	Text  & Image & Image + text   \\\midrule
BLUE  &    37.8  &  36.5   &   39 \\\midrule
METEOR   &    55.3   &  54  &  56.4 \\\midrule
    $\Delta \% \approx $ &   96\%  &     93\%   & 100\% \\\midrule
 
\bottomrule
\end{tabular}

    \caption{Performances of Multimodal Translation}
    \label{tab:M-MClass-Performances}
\end{table}

 \subsection{MultiModal  Movies classification  }
\label{ssec:MultiModal-Movies }
We use the same method in~\cite{arevalo2017gated } to extract text and image features and to get out our results  using the Word2Vec  model trained on the textual modal only  and using only  Image modality only with the VGG Network model to extract features and make predictions, finally, we  The Word2Vec model was trained on the textual modal alone, and only the Image modal was used with the VGG Network model to extract features and generate predictions. Finally, we utilize the same methodology as in \cite{arevalo2017gated} to pull out our findings.
\begin{table*}[t]
    \centering
    \begin{tabular}{L{1cm}L{2cm}L{1.8cm}L{1.5cm}}
\toprule

 Mertic &	Text (MLPW2V) & Image(VGG)	& MMU   \\\midrule
	Micro    &  0.595  & 0.437   &  0.630  \\\midrule
	Macro     & 0.488 & 0.284  &0.541 \\\midrule
             $\Delta \% \approx $        &   90 \% &     AVG 60.4  \%  &  100\%\\\midrule
 
\bottomrule
\end{tabular}
    \caption{Performances of  Multimodal Movie Classification}
    \label{tab:M-Translation}
\end{table*}
 \subsection{MultiModal  food classification  }
\label{ssec:MultiModal-Movies }
 
\begin{table*}[t]
    \centering
     \begin{tabular}{L{1cm}L{2cm}L{1.8cm}L{1.5cm}}
\toprule

 Mertic &	Text & Image	& Multimodal   \\\midrule
	Accuracy    & 84.41   & 	 71.67 & 92.50 \\\midrule

\bottomrule
\end{tabular}
    \caption{Performances of  Multimodal food Classification}
    \label{tab:M-food}
\end{table*}

 \section{Discussion}
 We observe with the experimental results using different datasets  in Tables \ref{tab:M-SA-Performances},\ref{tab:M-ER-Performances},\ref{tab:M-HS-Performances},\ref{tab:M-MClass-Performances},\ref{tab:M-food},\ref{tab:M-Translation} show 
 that multimodal class is outperforming the unimodal class for each dataset we used from 1\% to 30\% in terms of $ \Delta \% $ deviation accuracy metric for all sentiment and recognition analysis datasets in both aligned and not aligned dataset versions.
 In multimodal translation, we find that Multi modality improves with  4\% for Text and 7\% for image modality.
 in Hateful speech datasets, Text-Image model achieves  13\% to 20\% performance enhancement compared to Image or Text only models, except for the MMHS150K dataset where we get 2\%  Accuracy improvement because we measure the inter annotators without using the Cohen’s kappa score  instead we get a binary label from three annotators by only there is hate or not hat annotation in the three labels.    
 in the movie and food classification task, we find over 10\% for the text models and from 30\% to 45\% improvement than the  image models.

 On the other hand, and for the unimodal experiments, based on the aforementioned results, we observe that textual modality is the dominant modality in all tasks except in IEMOCAP the Acoustic modality is overcome the others, Table \ref{tab:modality-def}  shows the exact improvement percentages between pairs modalities over all datasets.

\begin{table*}[t]
    \centering
    		\begin{center}
			\begin{tabular}{|c|C{1.6cm} C{1.6cm}|c}
				\cline{2-3}
              
				\multicolumn{1}{c|}{ $\Delta \% $ } &  Text/Audio  & Text/video              \\\cline{1-3} 
               \multirow{2}{*}{ CMU-MOSEI} & 30 \%     &  27\% & A \\
                   &    15\% &   19 \% & UA  \\\cline{1-3}
               \multirow{2}{*}{CMU-MOSI } &  34.8  \%    &  62.11\%   & A   \\
                    &     43\%  &  35.8\%  & UA \\\cline{1-3}
                \multirow{2}{*}{ IEMOCAP} &    -3\%    & 3.1\% & A   \\
                    &  -0.4 \%  & 0 \%  & UA  \\\cline{1-3}
               \multirow{1}{*}{ UR-FUNNY} &   16  \%  &  17\%   \\		
               \multirow{1}{*}{ MMHS150K} &     -  &  18.6  \% \\ 	
               \multirow{1}{*}{ HATEMEMES} &      -&  5.8\%   \\ 	
               \multirow{1}{*}{ FOOD101} &    -  & 17 \%   \\ 	
               \multirow{1}{*}{ MM-IMB} &     - & 50 \%   \\	
                \multirow{1}{*}{Multi30k } &   -   &  3 \%  \\\cline{1-3}	
			\end{tabular}
		\end{center}
 
    \caption{Performances improvement for each modality pair [A] is for Aligned modality, [UA]- is for Unaligned modalities}
    \label{tab:modality-def}
\end{table*}

 The influence of modalities can be investigated  with two other methodologies 1) missing modality when our models attend to all modalities but for some real-word reason, we miss a modality.
 2) also if we get noise in our modalities this will affect the model performance.
 
 \section{Conclusion}
\label{sec:conclusions}

We empirically find that  modalities don't impact in the same way as a prediction in multimodal approaches. We evaluate SOTA models across multiple benchmark datasets on different tasks.
Based on the findings, unimodal performance is not worse than the multimodal  one, and we observe  a Textual modality can robust Transformer via multi-task optimization.
 We plan to explore the effectiveness of unimodal-based methods on the easiest problems, e.g., when the data samples can only predict from  one modality if possible.
 


\bibliography{My-references}
\bibliographystyle{alpha}

\end{document}